\documentclass[sigconf]{acmart}

\usepackage{multirow}
\usepackage{array}
\usepackage{color}

\AtBeginDocument{%
  \providecommand\BibTeX{{%
    \normalfont B\kern-0.5em{\scshape i\kern-0.25em b}\kern-0.8em\TeX}}}

\setcopyright{iw3c2w3}
\copyrightyear{2021}
\acmYear{2021}
\acmDOI{10.1145/3442381.3449895}

\acmConference[WWW '21]{Proceedings of the Web Conference 2021}{April 19--23, 2021}{Ljubljana, Slovenia}
\acmBooktitle{Proceedings of the Web Conference 2021 (WWW '21), April 19--23, 2021,
Ljubljana, Slovenia}
\acmPrice{}
\acmISBN{978-1-4503-8312-7/21/04}
\begin{document}

\title{A Trigger-Sense Memory Flow Framework for Joint Entity and Relation Extraction}


\author{Yongliang Shen}
\affiliation{%
  \institution{Zhejiang University}
  \country{}
}
\email{syl@zju.edu.cn}

\author{Xinyin Ma}
\affiliation{%
  \institution{Zhejiang University}
    \country{}
}
\email{maxinyin@zju.edu.cn}

\author{Yechun Tang}
\affiliation{%
  \institution{Zhejiang University}
    \country{}
}
\email{tangyechun@zju.edu.cn}

\author{Weiming Lu}
\affiliation{%
  \institution{Zhejiang University}
    \country{}
}
\email{luwm@zju.edu.cn}
\authornote{Corresponding author}

\renewcommand{\shortauthors}{Yongliang Shen, et al.}

\begin{abstract}

Joint entity and relation extraction framework constructs a unified model to perform entity recognition and relation extraction simultaneously, which can exploit the dependency between the two tasks to mitigate the error propagation problem suffered by the pipeline model. Current efforts on joint entity and relation extraction focus on enhancing the interaction between entity recognition and relation extraction through parameter sharing, joint decoding, or other ad-hoc tricks (e.g., modeled as a semi-Markov decision process, cast as a multi-round reading comprehension task). However, there are still two issues on the table. First, the interaction utilized by most methods is still weak and uni-directional, which is unable to model the mutual dependency between the two tasks. Second, relation triggers are ignored by most methods, which can help explain why humans would extract a relation in the sentence. They're essential for relation extraction but overlooked. To this end, we present a \textbf{Trigger-Sense Memory Flow Framework (TriMF)} for joint entity and relation extraction. We build a memory module to remember category representations learned in entity recognition and relation extraction tasks. And based on it, we design a multi-level memory flow attention mechanism to enhance the bi-directional interaction between entity recognition and relation extraction. Moreover, without any human annotations, our model can enhance relation trigger information in a sentence through a trigger sensor module, which improves the model performance and makes model predictions with better interpretation. Experiment results show that our proposed framework achieves state-of-the-art results by improves the relation F1 to 52.44\% (+3.2\%) on SciERC, 66.49\% (+4.9\%) on ACE05, 72.35\% (+0.6\%) on CoNLL04 and 80.66\% (+2.3\%) on ADE.

\end{abstract}

%
%




\maketitle

\section{Introduction}
Entity recognition and relation extraction aim to extract structured knowledge from unstructured text and hold a critical role in information extraction and knowledge base construction. For example, given the following text: \textit{ Ruby shot Oswald to death with the 0.38-caliber Colt Cobra revolver in the basement of Dallas City Jail on Nov. 24, 1963, two days after President Kennedy was assassinated.}, the goal is to recognize entities about \textit{ People}, \textit{ Location} and extract relations about \textit{ Kill}, \textit{Located in}  held between recognized entities. There are two things of interest to humans when carrying out this task. First, potential constraints between the relation type and the entity type, e.g., the head and tail entities of the \textit{Kill} are of \textit{People} type, and the tail entity of the \textit{Located in} is of \textit{Location} type. Second, triggers for relations, e.g. with words \textit{shot} and \textit{death}, the fact \textit{(Ruby, Kill, Oswald)} can be easily extracted from the above example.

Current entity recognition and relation extraction methods fall into two categories: pipeline methods and joint methods. Pipeline methods label entities in a sentence through an entity recognition model and then predict the relation between them through a relation extraction model \cite{chan2011exploiting, lin2016neural}. Although it is flexible to build pipeline methods, there are two common issues with these methods. First, they are more susceptible to error prorogation wherein prediction errors from entity recognition can affect relation extraction. Second, they lack effective interaction between entity recognition and relation extraction, ignoring the intrinsic connection and dependency between the two tasks. To address these issues, many joint entity and relation extraction methods are proposed and have achieved superior performance than traditional pipeline methods. In these methods, an entity recognition model and a relation extraction model are unified through different strategies, including constraint-based joint decoding \cite{li2014incremental,wang2018joint}, parameter sharing \cite{bekoulis2018joint,luan2018multi,eberts2019span}, cast as a reading comprehension task \cite{li2019entity, zhaoasking} or hierarchical reinforcement learning \cite{takanobu2019hierarchical}. Current joint extraction models have made great progress, but the following issues still remain:

\begin{enumerate}
    \item \textbf{Trigger information is underutilized in entity recognition and relation extraction.} Before neural information extraction models, rule-based entity recognition and relation extraction framework were widely used. They were devoted to mine hard template-based rules or soft feature-based rules from text and match them with instances  \cite{hearst1992automatic, jones1999bootstrapping, agichtein2000snowball, batista2015semi, aone1998sra, miller2000novel, fundel2007relex}. Such methods provide good explanations for the extraction work, but the formulation of rules requires domain expert knowledge or automatic discovery from a large corpus, suffering from tedious data processing and incomplete rule coverage.    End-to-end neural network methods have made great progress in the field of information extraction in recent years.  To exploit the rules, many works have begun to combine traditional rule-based methods by introducing a neural matching module \cite{zhou2020nero, lin2020triggerner,wang2019learning}. However, these methods still need to formulate seed rules or label seed relation triggers manually, and iteratively expand them.
    \item \textbf{The interaction between entity recognition and relation extraction is insufficient and uni-directional.} Entity recognition and relation extraction tasks are supposed to be mutually beneficial, but joint extraction methods do not take full advantage of dependency between the two tasks. Most joint extraction models are based on parameter sharing, where different task modules share input features or internal hidden layer states. However, these methods usually use independent decoding algorithms, resulting in a weak interaction between the entity recognition module and the relation extraction module. The joint decoding-based extraction model strengthens the interaction between modules, but it requires a trade-off between the richness of features for different tasks and joint decoding accuracy. Other joint extraction methods, such as modeling the task as a reading comprehension problem \cite{li2019entity, zhaoasking} or a semi-Markov process \cite{takanobu2019hierarchical},  still suffer from a lack of bi-directional interaction due to the sequential order of subtasks. 
    More specifically, if relation extraction follows entity recognition, the entity classification task will ignore the solution of the relation classification task. 
    \item \textbf{There is no distinction between the syntactic and semantic importance of words in a sentence.} We note that some words have a significant syntactic role but contribute little to the semantics of a sentence, such as prepositions and conjunctions. While some words are just the opposite, they contribute significantly to the semantics, such as nouns and notional verbs. When encoding context, most methods are too simple to inject syntactic features into the word vector, ignoring the fact that words differ in their semantic and syntactic importance. For example, some methods concatenate part of speech tags of words onto their semantic vectors via an embedding layer \cite{miwa2016end, fu2019graphrel}. Other methods combine the word, lexical, and entity class features of the nodes on the shortest entity path in the dependency tree to get the final features, which are then concatenated onto the semantic vector  \cite{bunescu2005shortest, miwa2016end}. These methods do not distinguish the two roles of a word for sentence semantics and syntax, but rather treat both roles of all words as equally important.
\end{enumerate}

In this paper, we propose a novel framework for joint entity and relation extraction to address the issues mentioned above. First, our model makes full use of relation triggers, which can indicate a specific type of relation. Without any relation trigger annotations, our model can extract relation triggers in a sentence and provide them as an explanation for model predictions. Second, to enhance the bi-directional interaction between entity recognition and relation extraction tasks, we design a Memory Flow Attention module. It stores the already learned entity category and relation category representations in memory. Then we adopt a memory flow attention mechanism to compute memory-aware sentence encoding, and make the two subtasks mutually boosted by enhancing task-related information of a sentence.
The Memory Flow Attention module can easily be extended to multiple language levels, enabling the interaction between the two subtasks at both subword-level and word-level. Finally, we distinguish the syntactic and semantic importance of a word in a sentence and propose a node-wise Graph Weighted Fusion module to dynamically fuse the syntactic and semantic information of words.

Our main contributions are as follow:

\begin{itemize}
    \item Considering the relation triggers, we propose the Trigger Sensor module, which implicitly extracts the relation triggers from a sentence and then aggregates the information of triggers into span-pair representation. Thus, it can improve the model performance and strengthens the model interpretability.
    \item To model the mutual dependency between entity recognition and relation extraction, we propose the Multi-level Memory Flow Attention module. This module constructs entity memory and relation memory to preserve the learned representations of entity and relation categories. Through the memory flow attention mechanism, it enables the bi-directional interaction between entity recognition and relation extraction tasks at multiple language levels.
    \item Since the importance of semantic and syntactic roles that words play in a sentence are different, we propose a node-wise Graph Weighted Fusion module to dynamically fuse semantic and syntactic information.
    \item Experiments show that our model achieves state-of-the-art performance consistently on the SciERC, ACE05, CoNLL04, and ADE datasets, and outperforms several competing baseline models on relation F1 score by 3.2\% on SciERC, 4.9\% on ACE05, 0.6\% on CoNLL04 and 2.3\% on ADE.
\end{itemize}

\section{Related Work}
\subsection{Rule-based Relation Extraction}

Traditional relation extraction methods utilize template-based rules \cite{aone1998sra, miller2000novel, fundel2007relex}, which are first formulated by domain experts or automatically generated from a large corpus based on statistical methods. Then, they apply hard matching to extract the corresponding relation facts corresponding to the rules. Later on, some works change the template-based rules to feature-based rules (such as TF-IDF, CBOW) and extract relations by soft matching \cite{kambhatla2004combining, zhang2006exploring, jiang2007systematic, bui2011hybrid}, but still could not avoid mining the rule features from a large corpus using statistical methods. In short, rule-based relation extraction models typically suffer from a number of disadvantages, including tedious efforts on the rule formulation, a lack of extensibility, and low accuracy due to incomplete rule coverage, but they can provide a new idea for neural relation extraction systems.

Some recent efforts on neural extraction systems attempt to focus on rules or natural language explanations \cite{wang2019learning}. NERO \cite{zhou2020nero} explicitly exploits labeling rules over unmatched sentences as supervision for training RE models. It consists of a sentence-level relation classifier and a soft rule matcher. The former learns the neural representations of sentences and classifies which relation it talks about. The latter is a learnable module that produces matching scores for unmatched sentences with collected rules. NERO labels sentences according to predefined rules, and makes full use of information from unmatched instances. However, it is still a tedious process to formulate seed rules manually. And the quality of rule-making affects the performance of the entire system.

\subsection{Joint Entity and Relation Extraction}

Previous entity and relation extraction models are pipelined \cite{chan2011exploiting, lin2016neural}. In these methods, an entity recognition model first recognizes entities of interest, and a relation extraction model then predicts the relation type between the recognized entities. Although pipeline models have the flexibility of integrating different model structures and learning algorithms, they suffer significantly from error propagation. To tackle this issue, joint learning models have been proposed. They fall into two main categories: parameter sharing and joint decoding methods.

Most methods jointly model the two tasks through parameter sharing \cite{miwa2016end, zheng2017joint}. They unite entity recognition and relation extraction modules by sharing input features or internal hidden layer states. Specifically, these methods use the same encoder to provide sentence encoding for both the entity recognition module and the relation extraction module. Some methods  \cite{bekoulis2018adversarial, luan2018multi, luan2019general, wadden2019entity} perform entity recognition first and then pair entities of interest for relation classification. While other methods  \cite{takanobu2019hierarchical, yuan2020relation} are the opposite, they predict possible relations first and then recognize the entities in the sentence.
DygIE \cite{luan2019general} constructs a span-graph and uses message propagation methods to enhance interaction between entity recognition and relation extraction. HRL \cite{takanobu2019hierarchical} models the joint extraction problem as a semi-Markov decision process, and uses hierarchical reinforcement learning to extract entities and relations.
CASREL \cite{wei2020novel} considers the general relation classification as a tagging task. Each relation corresponds to a tagger that recognizes the tail entities based on a head entity and context.
CopyMTL \cite{zeng2018extracting} casts the extraction task as a generation task and proposes an encoder-decoder model with a copy mechanism to extract relation tuples with overlapping entities.
Although entity recognition and relation extraction modules can adopt different structures in these methods, their independent decoding algorithms result in insufficient interaction between the two modules. Furthermore, subtasks are performed sequentially in these methods, so the interaction between two tasks is uni-directional.

To enhance the bi-directional interaction between entity recognition and relation extraction tasks, some joint decoding algorithms have been proposed. 
\cite{yang2013joint} proposes to use integer linear planning to enforce constraints on the prediction results of the entity and relation models.   
\cite{katiyar2016investigating} uses conditional random fields for both entity and relation models and obtains the output results of the entity and relation by the Viterbi decoding algorithm. Although the joint decoding-based extraction model strengthens the interaction between two modules, it still requires a trade-off between the richness of features required for different tasks and the accuracy of joint decoding.

\section{Trigger-Sense Memory Flow Framework}

\subsection{Framework Overview}

In this section, we will introduce the \textbf{Trigger-Sense Memory Flow Framework (TriMF)} for joint entity and relation extraction, which consists of five main modules: \textbf{Memory} module,  \textbf{Multi-Level Memory Flow Attention} module, \textbf{Syntactic-Semantic Graph Weighted Fusion} module, \textbf{Trigger Sensor} module, and \textbf{Memory-Aware Classifier} module.

\begin{figure*}[h]
  \centering
  \includegraphics[width=\linewidth]{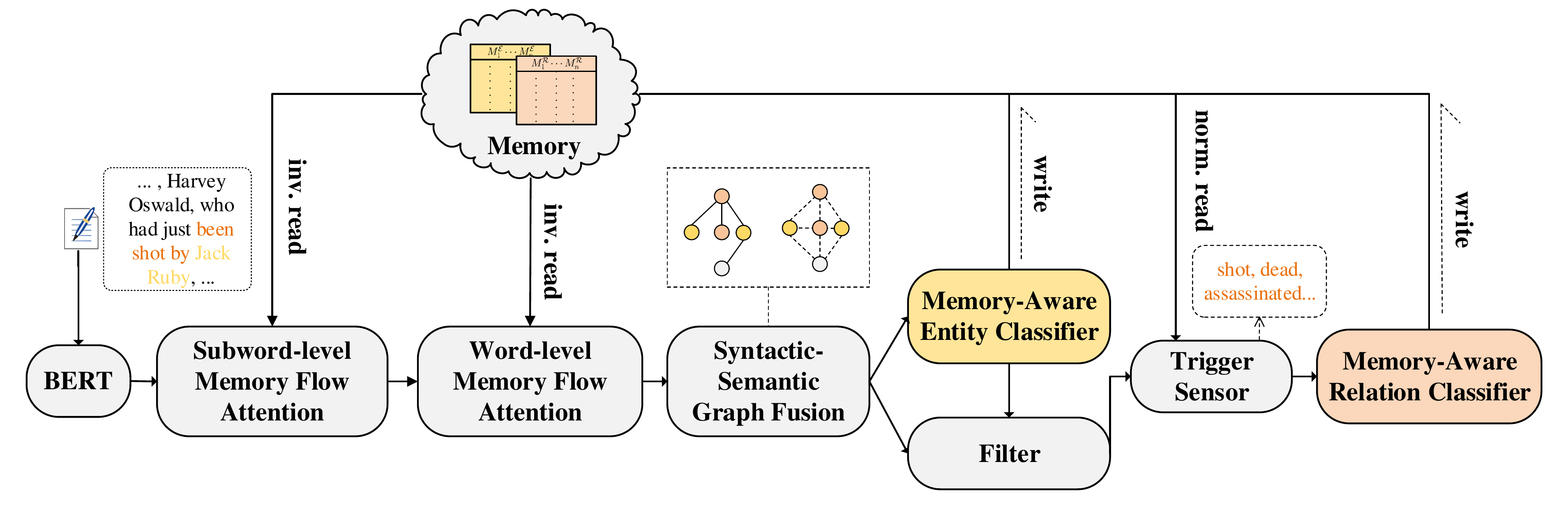}
  \caption{Trigger-Sense Memory Flow Framework (TriMF) Overview}
   \label{fig:overview}
\end{figure*}

The overall architecture of the TriMF is illustrated in Figure \ref{fig:overview}.  We first initialize the Memory, including an Entity Memory  $\mathbf{M}^\mathcal{E} \in \mathbb{R}^{n^e \times h_{me}} $ and a Relation Memory  $\mathbf{M}^\mathcal{R}  \in  \mathbb{R}^{n^r\times h_{mr}} $, where $n^e$ and $n^r$ denote the number of entity categories and relation categories, $h_{me}$ and $h_{mr}$ denote the slot size of entity memory and the relation memory.

\begin{figure}[h]
  \centering
  \includegraphics[width=\linewidth]{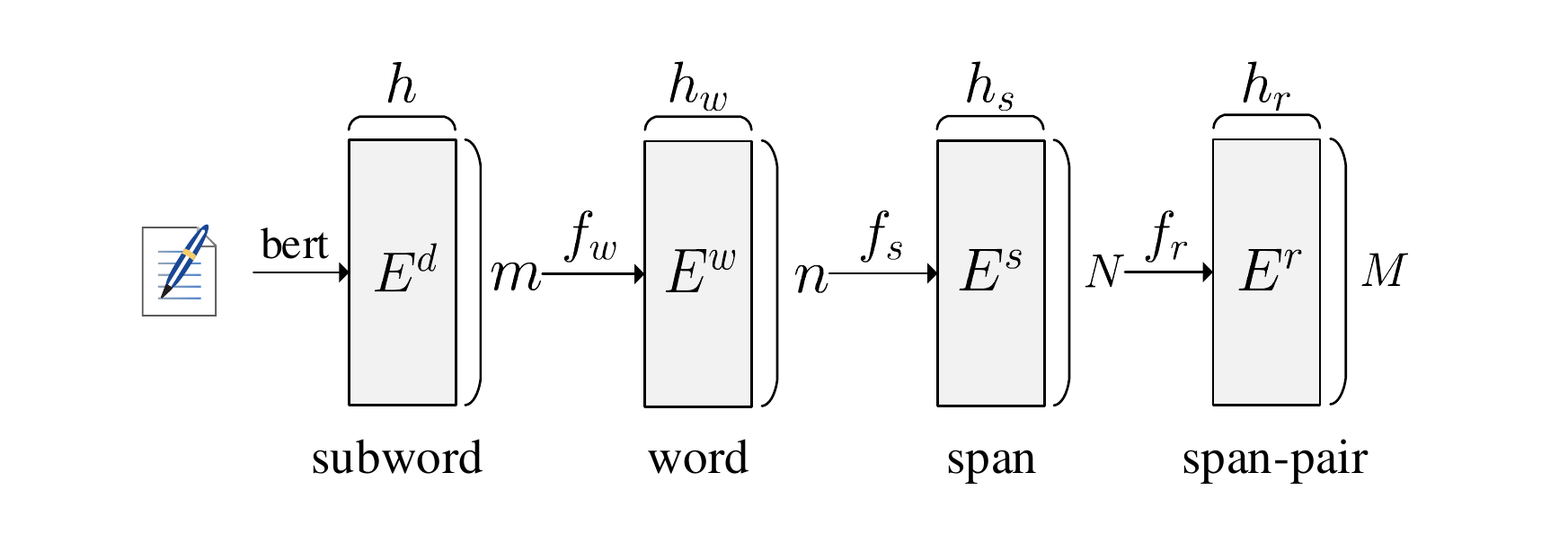}
  \caption{ Four Levels Encoding}
   \label{fig:encode}
\end{figure}

Our model performs a four-level sentence encoding (subword, word, span, and span-pair,  as shown in Figure \ref{fig:encode}) and two-step classification (entity classification and relation classification). More specifically, a sentence is encoded by BERT \cite{devlin2018bert} to obtain subword sequence encoding $\mathbf{E}^d=\mathbb{R}^{m\times h}$, where $m$ denotes the number of subwords in the sentence, and $h$ denotes the hidden state size of BERT. 
Based on $\mathbf{M}^\mathcal{R}$, $\mathbf{M}^\mathcal{E} $ and  $\mathbf{E}^d$, we perform the first Memory Flow Attention at the subword-level.
Then we use $f_w$ to aggregate the subword sequence encoding into a word sequence encoding $\mathbf{E}^w=\mathbb{R}^{n \times h_w}$, where $n$ denotes the number of words in the sentence, and $h_w$ denotes the size of the word vector. Here for $f_w$, we adopt the max-pooling function. 
Based on $\mathbf{M}^\mathcal{R} $, $\mathbf{M}^\mathcal{E} $ and  $\mathbf{E}^w$, we perform the second  Memory Flow Attention at the word-level. After that, the word sequence encoding is fed into the Syntactic-Semantic Graph Weighted Fusion module to fuse semantic and syntactic information at the word-level.
Then, we combine the word sequence encodings by $f_s$ to obtain the span sequence encodings $\mathbf{E}^s=\mathbb{R}^{N \times h_s} $, where $N$ denotes the number of spans in the sentence, and $h_s$ denotes the size of the span vector. Here for $f_s$, we adopt a method of concatenating a span-size embedding on max-pooled word embeddings. We filter out the spans which are classified as the \textit{None} category by a Memory-Aware Entity Classifier. After pairing the spans of interest, We compute local-context representation $g_{local}$ and full-contextual span-pair specific trigger representation $\mathbf{g}_{trigger}$ using the Trigger Sensor. We combine the encodings of the head span, tail span, $g_{local}$ and $\mathbf{g}_{trigger}$ to obtain the encoding $\mathbf{E}^r  \in \mathbb{R}^{M\times h_r}$, where $\mathbf{E}^r_{\left(ij\right)}$ denotes the span pair encoding consisting of the $i^{th}$ and $j^{th}$ spans, $M$ denotes the number of candidate span pairs, and $h_r$ denotes the size of the span pair encoding. 
Lastly, we input the candidate span-pair representation to the Memory-Aware Relation Classifier and predict the relation type between the two spans.
In the next sections, we'll cover five main modules of our model in detail.

\subsection{Memory} 

Memory holds category representations learned from historical training examples, consist of entity memory and relation memory. Each slot of these two memories indicates an entity category and a relation category respectively. The category representation is held in the corresponding memory slot, which can be used by the Memory Flow Attention module to enhance information related to the tasks in a sentence, or by the Trigger Sensor module to sense triggers. 

In the Memory module, we define two types of processes, Memory Read Process and Memory Write Process, to manipulate the memory.

\noindent \textbf{Memory Read Process} Given an input $\mathbf{E}$ and our memory $\mathbf{M}$, we define two processes to read memory: \textit{normal read process} and \textit{inverse read process}. The normal read process takes the input as \textit{query}, the memory as \textit{key} and \textit{value}. First, we calculate the attention weights of the input $\mathbf{E}$ on the memory $\mathbf{M}$ by bilinear similarity function, and then we weigh the memory by the weights.


\begin{equation}
    \operatorname{A}_{norm}\left(\mathbf{E},\mathbf{M}\right)=\operatorname{softmax} \left(  \mathbf{E}\mathbf{W}\mathbf{M}^T\right)  
\end{equation}

\begin{equation}
        \operatorname{Read}_{norm}\left(\mathbf{E},\mathbf{M}\right) =\operatorname{A}_{norm}\left(\mathbf{E},\mathbf{M}\right)\mathbf{M}
\end{equation}

\noindent where $\mathbf{W}$ is a learnable parameter for the bilinear attention mechanism. While the inverse read process takes the memory as \textit{query}, the input as \textit{key} and \textit{value}. We first compute 2d-attention weight matrix through bilinear similarity function, and then sum the 2d-attention weight matrix on the memory-slot dimension to obtain a 1d-attention weight vector on the input $\mathbf{E}$. The more relevant element in input with the memory has a larger weight. We then multiply the 1d-attention weight vector with $\mathbf{E}$ to get a memory-aware sequence encoding:


\begin{equation}
    \operatorname{A}_{inv}\left(\mathbf{E},\mathbf{M}\right)=\sum\limits_{i=1}^{|\mathbf{M}|} \operatorname{softmax} \left(\mathbf{M}_i \mathbf{W}\mathbf{E}^T\right)  
\end{equation}

\begin{equation}
    \operatorname{Read}_{inv}(\mathbf{E},\mathbf{M} ) =\operatorname{A}_{inv}\left(\mathbf{E},\mathbf{M}\right)\mathbf{E}
\end{equation}

\noindent where $\mathbf{W}$ is a learnable parameter for the bilinear attention mechanism and $|\mathbf{M}|$ denotes the number of slots in the memory $\mathbf{M}$.

\noindent \textbf{Memory Write Process} We write entity memory using gradients of entity classification losses and write relation memory using gradients of relation classification losses. If the gradient of the current instance's classification loss is large, it means that the classified instance (span or span-pair) representation is far away from the corresponding memory slot (entity or relation category representation of ground truth) while closer to the memory slots of the other categories, and we need to assign a large weight to this instance when writing it into memory. This makes the representations of the categories stored in memory more accurate. The write process for entity memory and relation memory is described below:

\begin{equation}
    \mathbf{M}^\mathcal{E}_e=  \mathbf{M}^\mathcal{E}_e- \mathbf{E}^s_i \mathbf{W}^e   \frac{\partial \mathcal{L}^e}{\partial logit_e}    lr
\end{equation}

\begin{equation}
    \mathbf{M}^\mathcal{R}_r= \mathbf{M}^\mathcal{R}_r- \mathbf{E}^r_{(ij)}  \mathbf{W}^r  \frac{\partial \mathcal{L}^r}{\partial logit_r}  lr
\end{equation}

\begin{equation}
    logit_e=log\left(\frac{p(s_i=e)}{1-p(s_i=e)}\right)
\end{equation}

\begin{equation}
    logit_r=log\left(\frac{p(r_{ij}=r)}{1-p(r_{ij}=r)}\right)
\end{equation}

\noindent where $\mathcal{L}^e$ and $\mathcal{L}^r$ denote entity classification loss and relation classification loss,  $lr$ denotes the learning rate, $\mathbf{W}^e$ and $\mathbf{W}^r$ are two weight matrices, $p(s_i=e)$ denotes the probability of span $s_i$ belonging to entity type $e$, $p(r_{ij}=r)$ denotes the probability of span-pair's relation $r_{ij}$ belonging to relation type $r$,  and $\mathbf{E}^s_i$, $\mathbf{E}^r_{ij}$ denote candidate span and span-pair encoding, respectively. The above symbols are specifically defined in defined at Sec.\ref{sec:ent_classifier}.

\subsection{Multi-level Memory Flow Attention}

We perform a memory flow attention mechanism between the memory and the input sequence to enhance task-relevant information, such as entity surface names and trigger words. Entity memory and relation memory can enhance entity-related and relation-related information in the input instance for the two tasks respectively, thus they can help to strengthen bi-directional interaction between tasks.

\noindent\textbf{Memory Flow Attention} 
In order to enhance the task-relevant information in a sentence, we designed the Memory Flow Attention based on the Memory. Given a memory $\mathbf{M}$ and a sequence encoding $\mathbf{E}$, We calculate the memory-aware sequence encoding by runing \textit{memory inverse read process}:

\begin{equation}
    \operatorname{MFA}_{s}\left(\mathbf{E},\mathbf{M} \right) =\operatorname{Read}_{inv}\left(\mathbf{E},\mathbf{M}\right)
\end{equation}

A single memory flow can be extended to multiple memory flows. We consider two types in our work: relation memory flow and entity memory flow. So we design a Multi-Memory Flow Attention mechanism,  which is calculated as follows:

\begin{equation}
    \operatorname{MFA}_{m}(\mathbf{E},\mathbf{M}^\mathcal{R} ,\mathbf{M}^\mathcal{E} ) =\operatorname{mean}\left( \operatorname{MFA}_{s}\left(\mathbf{E},\mathbf{M}^\mathcal{R} \right), \operatorname{MFA}_{s}\left(\mathbf{E},\mathbf{M}^\mathcal{E} \right)\right)
\end{equation}

\noindent where $\mathbf{M}^\mathcal{E}$ and  $\mathbf{M}^\mathcal{R}$ denote entity and relation memory respectively. we know that languages are hierarchical, and different levels represent semantic information at different levels of granularity. As shown in Figure \ref{fig:mfa}, we extend the multi-memory flow attention mechanism to multiple levels ( subword-level and word-level ), and design a Multi-Level Multi-Memory Flow Attention mechanism:

\begin{figure}[h]
  \centering
  \includegraphics[width=1.0\linewidth]{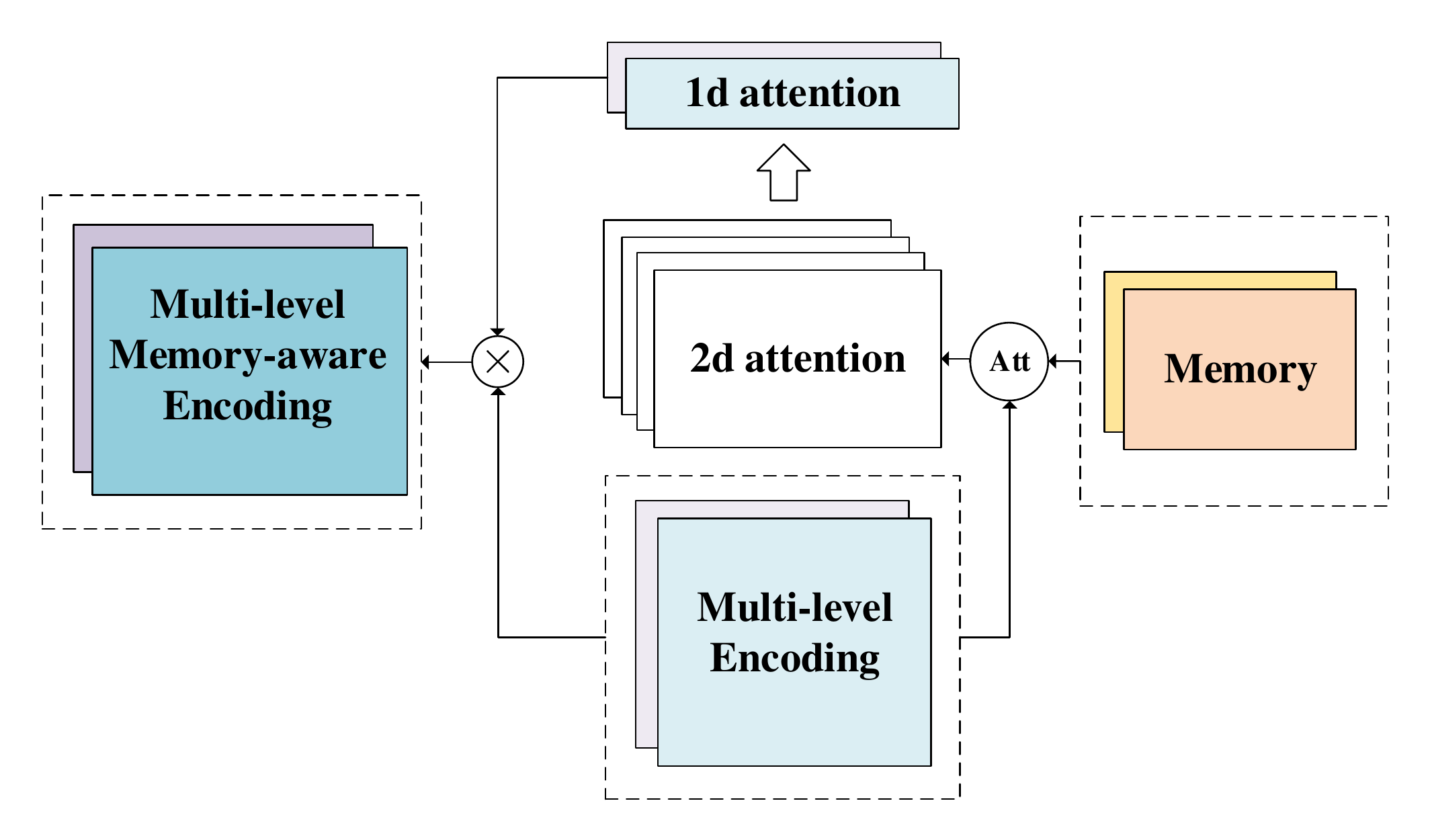}
  \caption{Multi-Level Multi-Memory Flow Attention}
   \label{fig:mfa}
\end{figure}

\begin{equation}
    \overline{\mathbf{E}}^d=\operatorname{MFA}_{m}(\mathbf{E}^d,\mathbf{M}^r,\mathbf{M}^e)  
\end{equation}

\begin{equation}
    \mathbf{E}^w=f_w\left(\mathbf{\overline{E}}^d\right)  
\end{equation}

\begin{equation}
    \overline{\mathbf{E}}^w=\operatorname{MFA}_{m}(\mathbf{E}^w,\mathbf{M}^r,\mathbf{M}^e)  
\end{equation}

\noindent where $\mathbf{\overline{E}}^d$ and  $\mathbf{\overline{E}}^w$ denote memory-aware sequence encoding at subword-level and word-level respectively. 

\subsection{Syntactic-Semantic Graph Weighted Fusion}

The semantic information and syntactic structure of a sentence are important for both entity recognition and relation extraction. We consider both by constructing semantic and syntactic graphs from a sentence, with nodes in the graph refer to words in the sentence. We update a node representation based on its neighbor nodes' representations and the graph structure in the two graphs. We note that some words have a significant syntactic role but contribute little to the semantics of a sentence, such as prepositions and conjunctions. While some words are just the opposite, they contribute significantly to the semantics, such as nouns and notional verbs. Therefore, we need to fuse syntactic and semantic graphs based on the relative importance of the syntactic role and semantic role. First, the nodes in the two graphs are initialized as: 

\begin{equation}
\mathbf{H}^{(0)} = \overline{\mathbf{E}}^w
\end{equation}

\noindent\textbf{Syntactic Graph}
We construct a directed syntactic graph from a sentence based on dependency parsing, with the word as a node and the dependency between words as an edge. We then use the R-GCN \cite{schlichtkrull2018modeling} to update node representations. The node representations of the syntactic graph $\widehat{\mathbf{H}}^{(l)}$ in $l^{th}$ layer are calculated as:

\begin{equation}
    \widehat{\mathbf{H}}_{i}^{(l)}=\sigma\left(\sum_{r \in \mathcal{R}_{dep}} \sum_{j \in \mathcal{N}_{i}^{r}} \frac{1}{c_{i, r}} \mathbf{\widehat{W}}_{r}^{(l)} \mathbf{H}_{j}^{(l)}+\mathbf{\widehat{W}}_{0}^{(l)} \mathbf{H}_{i}^{(l)}\right) 
\end{equation}

\noindent where $\mathbf{\widehat{W}}_{r}^{(l)}$ and $\mathbf{\widehat{W}}_{0}^{(l)}$ denote two learnable weight matrices, and $\mathcal{N}_{i}^{r}$ denotes the set of neighbor indices of node $i$ under relation $r \in \mathcal{R}_{dep}$.

\noindent\textbf{Semantic Graph}
We compute the dense adjacency matrix based on semantic similarity and randomly sample from the fully connected graph to construct the semantic graph:

\begin{equation}
    \mathbf{\alpha}=\operatorname{LeakyReLU}\left(\mathbf{\widetilde{W}} \mathbf{H}^{(l)}\right)^{T} \operatorname{LeakyReLU}\left(\mathbf{\widetilde{W}} \mathbf{H}^{(l)}\right)
\end{equation}

\noindent where $\mathbf{\widetilde{W}}$ denotes a trainable weight matrix. Then we compute a weighted average for aggregation of neighbor nodes $\mathcal{N}(i)$, where the weights come from the normalized adjacency matrices $\mathbf{\overline{\alpha}}$. We update the node representations of semantic graph $ \widetilde{\mathbf{H}}_{i}^{(l)}$ in $l^{th}$ layer, which are calculated as follows:

\begin{equation}
    \overline{\mathbf{\alpha}}=\operatorname{softmax\left(\mathbf{\alpha}\right)}
\end{equation}

\begin{equation}
    \widetilde{\mathbf{H}}_{i}^{(l)}=\overline{\alpha}_{i, i} \mathbf{\widetilde{W}} \mathbf{H}_{i}^{(l)}+\sum_{j \in \mathcal{N}(i)} \overline{\alpha}_{i, j} \mathbf{\widetilde{W}} \mathbf{H}_{j}^{(l)} 
\end{equation}


\noindent\textbf{Node-Wise Graph Weighted Fusion} We design a graph weighted fusion module to dynamically fuse two graphs according to the relative semantic and syntactic importance of words in a sentence. The [CLS] vector, denote as $\mathbf{e}^{cls} $, is often used for sentence-level tasks and contains information about the entire sentence. We first calculate the bilinear similarity between $\mathbf{e}^{cls} $ and each node of semantic and syntactic graphs. Then we normalize the similarity vectors across two graphs to obtain two sets of weights, which indicate semantic and syntactic importance respectively. Finally, we fuse all nodes across the graphs based on the weights:

\begin{equation}
    \mathbf{\overline{w}},\mathbf{\widehat{w}}=\operatorname{softmax}\left(\left\{\mathbf{e}^{cls} \mathbf{W} \widetilde{\mathbf{H}}^{(l)} ,\mathbf{e}^{cls}\mathbf{W} \widehat{\mathbf{H}}^{(l)}\right\}\right)
\end{equation}

\begin{equation}
\mathbf{H}^{(l+1)}_i =\mathbf{\widetilde{w}}_i \cdot \widetilde{\mathbf{H}}^{(l)}_i + \mathbf{\widehat{w}}_i\cdot \widehat{\mathbf{H}}^{(l)}_i 
\end{equation}

\noindent where $\mathbf{W}$ is a learnable weight matrix, $\mathbf{\widetilde{w}}$ and $\mathbf{\widehat{w}}$ denote the node importance weights of syntactic and semantic graphs, respectively. Then we map the node representations $\mathbf{H}^{(l+1)}$ to the corresponding word representations $E^g$ using mean-pooling:

\begin{equation}
\mathbf{E}^g = \operatorname{mean} \left(\mathbf{H}^{(l+1)},\mathbf{\overline{E}}^w\right)
\end{equation}

\subsection{Trigger Sensor}

We know that a particular relation usually occurs in conjunction with a particular set of words, which we call relation triggers. They can help explain why humans would extract a relation in the sentence and play an essential role in relation extraction. 
We present a Trigger Sensor module that senses and enhances the contextual trigger information without any trigger annotations.

Relation triggers typically appear in local context between a pair of spans $\left(s_{i}, s_{j}\right) $, and some approaches encode local context directly into the span-pair representation for relation classification. However, these approaches do not consider the case where the triggers are outside the span-pair, resulting in the model ignoring useful information from other contexts. We design both a Local-Context Encoder and a Full-Context Trigger Sensor to compute the local-context representation $\mathbf{g}_{local} $ and the full-context trigger representation $\mathbf{g}_{trigger} $. 

\noindent\textbf{Local-Context Encoder}
We aggregate local-context information between spans of interest using max-pooling. The local-context representation $\mathbf{g}_{local}$ is calculated as:

\begin{equation}
    \mathbf{g}_{local} =\max \left(\mathbf{E}^g_k, \mathbf{E}^g_{k+1},\cdots,\mathbf{E}^g_{h}\right)
\end{equation}
\noindent where $\mathbf{E^g_k}, \mathbf{E^g_{k+1}},\cdots,\mathbf{E^g_{h}}$ are the encodings of words between the two spans $\left(s_{i}, s_{j}\right) $.

\noindent\textbf{Full-Context Trigger Sensor}
Full-context trigger sensor aims to sense and enhance span-pair specific triggers. 
Given a pair of spans $\left(s_{i}, s_{j}\right) $, we use head span and tail span as queries respectively and execute \textit{normal read process} on the relation memory. After obtaining two span-specific memory representations, we perform mean-pooling across them to get the span-pair specific relation representation $m^r_{(ij)}$:

\begin{equation}
    m^r_{(ij)}=\operatorname{mean}\left(
    \operatorname{Read}_{norm}\left(\mathbf{E}^s_{i},\mathbf{M}^\mathcal{R}\right),
    \operatorname{Read}_{norm}\left(\mathbf{E}^s_{j},\mathbf{M}^\mathcal{R}\right)
    \right)
\end{equation}

We calculate the similarity between $\mathbf{m}^r_{(ij)}$ and each word representation of a word sequence, and then weigh the word sequence to get the full-context trigger representation $\mathbf{g}_{trigger} $.

\begin{equation}
    \textbf{g}_{trigger} =\operatorname{softmax}\left({\mathbf{m}_{(ij)}(\mathbf{E}^g)^T}\right)\mathbf{E}^g
\end{equation}

We incorporate the local-context representation $\mathbf{g}_{local}$ and the full-context trigger representation $\mathbf{g}_{trigger} $ into the span-pair encoding $\mathbf{E}^r_{ij}$ using $f_r$:

\begin{equation}
    \mathbf{E}^r_{ij}=f_r\left( \mathbf{E}^s_i,\mathbf{E}^s_j,\mathbf{g}_{local},\mathbf{g}_{trigger} \right) 
\end{equation}

\noindent for $f_r$ we adopt the concatenate function.

\noindent\textbf{Trigger Extraction}
Using the trigger sensor, we can also extract relation triggers and provide a reasonable explanation for model predictions. Based on the similarity of each word representation with the span-pair specific relation representations $m^r_{(ij)}$, we rank the words. The top-ranked words can be used as relation triggers to explain the model's predictions. We will show the trigger extraction ability of our model in the case study section. 

\subsection{Memory-Aware Classifier}
\label{sec:ent_classifier}
Representations of the entity and relation categories are stored in entity memory and relation memory, respectively. Based on the bilinear similarity between instance (span or span-pair) representation and categories representations,
we compute the probability of candidate span $s_i $ being an entity $e$:

\begin{equation}
    p\left(s_i=e\right)=\frac{\exp\left({\mathbf{E}^s_i \mathbf{W}^e  M^\mathcal{\mathbf{E}}_e }\right)}{\sum_{k \in \mathcal{E} }\exp\left({\mathbf{E}^s_i \mathbf{W}^e  \mathbf{M}^\mathcal{\mathbf{E}}_k }\right)}
\end{equation}
\noindent and the probability of candidate span-pair $\left(s_{i}, s_{j}\right) $ having a relation $r$:

\begin{equation}
    p\left(r_{(ij)}=r\right)=\operatorname{sigmoid}\left({\mathbf{E}^r_{(ij)} \mathbf{W}^r  \mathbf{M}^\mathcal{R}_r }\right)
\end{equation}

\noindent where $\mathbf{W}^e\in \mathbb{R}^{h_s\times h_{me}}$ and $\mathbf{W}^r\in \mathbb{R}^{h_r\times h_{mr}}$ denote two learnable weight matrices. Finally, we define a joint loss function for entity classification and relation classification:
$$\mathcal{L}=\mathcal{L}^{s}+\mathcal{L}^{r}$$

\noindent where $\mathcal{L}^{s} $ denotes the cross-entropy loss over entity categories(including the \textit{None} category), and $\mathcal{L}^{r} $ denotes the binary cross-entropy loss over relation categories.

\subsection{Two-Stage Training}

At the start of training, since the memory is randomly initialized, the Memory Flow Attention module and Trigger Sensor module will introduce noises to the sequence encoding. These noises further corrupt the semantic information of the pre-trained BERT \cite{devlin2018bert} through the gradient descent. We therefore divide the model training procedure into two stages. In the first stage, we aim to learn more accurate category representations and store them into the corresponding memory slots. We only train Memory-Aware Classifier and Graph Weighted Fusion modules and update the memory through the \textit{memory write process}. In the second stage, we add Memory Flow Attention and Trigger Sensor modules to the training procedure. Based on the more accurate representations of the categories stored in the memory, we can strengthen the contextual task-related features and relation triggers through \textit{memory read process}.

\section{Experiments}

\begin{table*}[]
\small
\begin{tabular}{@{}llccccccc@{}}
\toprule
  \multirow{2}{*}{Dataset} & \multirow{2}{*}{Model}  & \multicolumn{3}{c}{Entity} & \multicolumn{3}{c}{Relation} \\ 
 \cmidrule(lr){3-5} \cmidrule(lr){6-8}
  &  & Precision & Recall & F1 & Precision & Recall & F1 \\ \midrule
 \multirow{5}{*}{SciERC} & SciIE$\dagger$ & 67.20 & 61.50 & 64.20 & 47.60 & 33.50 & 39.30  \\
  & DyGIE$\dagger$ & - & - & 65.20 & - & - & 41.60 \\
  & DYGIE++$\dagger$ & - & - & 67.50 & - & - & 48.40  \\
  & SpERT$\dagger$ (using SciBERT  \cite{beltagy2019scibert}) & 70.87 & 69.79 & 70.33 & 53.40 & 48.54 & 50.84  \\
  & TriMF$\dagger$ (using SciBERT) & 70.18 ($\pm$0.65) & 70.17 ($\pm$0.94) & 70.17 ($\pm$0.56) & 52.63 ($\pm$1.24) & 52.32 ($\pm$1.73) & \textbf{52.44 ($\pm$0.40)}   \\\midrule 
 \multirow{6}{*}{ACE05} & DyGIE$\dagger$ & - & - & 88.40 & - & - & 63.20 \\
  & DYGIE++$\dagger$ & - & - & 88.60 & - & - & 63.40 \\
  & TriMF$\dagger$  &87.67 ($\pm$0.17) & 87.54 ($\pm$0.29) & 87.61 ($\pm$0.21) & 65.87 ($\pm$0.55) & 67.12 ($\pm$0.63)& \textbf{66.49 ($\pm$0.32)} \\ 
    \cmidrule(lr){2-8} 
  & Multi-turn QA$\ddagger$ & 84.70 & 84.90 & 84.80 & 64.80 & 56.2 & 60.20 \\
  & MRC4ERE++$\ddagger$ & 85.90 & 85.20 & 85.50 & 62.00 & 62.20 & 62.10 \\
   & TriMF$\ddagger$ &87.67 ($\pm$0.17) & 87.54 ($\pm$0.29) & \textbf{87.61 ($\pm$0.21)}  & 62.19 ($\pm$0.52) & 63.37 ($\pm$0.52) & \textbf{62.77 ($\pm$0.22)} \\ \midrule  
 \multirow{5}{*}{CoNLL04} & Multi-head + AT  \cite{bekoulis2018adversarial} $\ddagger$ &  &  &  83.9 &  &  & 62.04 \\
  & Multi-turn QA$\ddagger$ & 89.00 & 86.60 & 87.80 & 69.20 & 68.20 & 68.90 \\
  & SpERT$\ddagger$ & 88.25 & 89.64 & 88.94 & 73.04 & 70.00 & 71.47 \\
  & MRC4ERE++$\ddagger$ & 89.30 & 88.50 & 88.90 & 72.20 & 71.50 & 71.90 \\
  & TriMF$\ddagger$ & 90.26 ($\pm$0.62) & 90.34 ($\pm$0.60) & \textbf{90.30 ($\pm$0.24)} & 73.01 ($\pm$0.21) & 71.63 ($\pm$0.26) & \textbf{72.35 ($\pm$0.23)} \\ \midrule 
 \multirow{3}{*}{ADE}& Multi-head + AT  \cite{bekoulis2018adversarial} $\ddagger$* & - & - & 86.73 & - & - & 75.52 \\
  & SpERT$\ddagger$* & 88.99 & 89.59 & 89.28 & 77.77 & 79.96 & 78.84 \\
  & TriMF$\ddagger$* & 89.50 & 91.29 & \textbf{90.38} & 74.22 & 83.43 & \textbf{80.66} \\ \bottomrule
\end{tabular}
\caption{Precision, Recall, and F1 scores on the SciERC, ACE05, CoNLL04 and ADE datasets. (macro-average=*, boundary evaluation=$\dagger$, strict evaluation=$\ddagger$)}
\label{tab:results}
\end{table*}

\subsection{Datasets}

We evaluate TriMF described above using the following four datasets:

\begin{itemize}
\item \textbf{SciERC}: The SciERC  \cite{luan2018multi} includes annotations for scientific entities, their relations, and coreference clusters for 500 scientific abstracts. The dataset defines 6 types for annotating scientific entities and 7 relation categories. We adopt the same data splits as in \cite{luan2018multi}.
\item \textbf{ACE05}: ACE05 was built upon ACE04, and is commonly used to benchmark NER and RE methods. ACE05 defines 7 entity categories. For each pair of entities, it defines 6 relation categories. We adopt the same data splits as in \cite{miwa2016end}.

\item \textbf{CoNLL04}: The CoNLL04 dataset  \cite{roth2004linear} consists of 1,441 sentences with annotated entities and relations extracted from news articles. It defines 4 entity categories and 5 relation categorie. We adopt the same data splits as in \cite{gupta2016table}, which contains 910 training, 243 dev, and 288 test sentences.

\item \textbf{ADE}: The Adverse Drug Events (ADE) dataset  \cite{gurulingappa2012development} consists of 4, 272 sentences and 6, 821 relations extracted from medical reports. These sentences describe the adverse effects arising from drug use. ADE dataset contains two entity categories and a single relation category. 
\end{itemize}

\subsection{Compared Methods}
Our model is compared with current advanced joint entity and relation extraction models, divided into three types: general parameter-sharing based models (Multi-head AT, SPtree, SpERT, SciIE), span-graph based models (DyGIE, DyGIE++), and reading-comprehension based models (multi-turn QA, MRC4ERE).

\noindent \textbf{Multi-head + AT} \cite{bekoulis2018adversarial} treats the relation extraction task as a multi-head selection problem. Each entity is combined with all other entities to form entity pairs that can be predicted which relations to have. In addition, instead of being a multi-category task where each category is mutually exclusive, the relation classification is treated as multiple bicategorical tasks where each relation is independent, which allows more than one relation to be predicted.

\noindent \textbf{SPTree} \cite{miwa2016end}
shares parameters of the encoder in joint entity recognition and relation extraction tasks, which strengthens the correlation between the two tasks. SPTree is the first model that adopts a neural network to solve a joint extraction task for entities and relations.

\noindent \textbf{SpERT} \cite{eberts2019span} is a simple and effective model for joint entity and relation extraction. It uses BERT \cite{devlin2018bert} to encode a sentence, and enumerates all spans in the sentence. Then it performs span classification and span-pair classification to extract entities and relations.

\noindent \textbf{SciIE} \cite{luan2018multi}
is a framework for extracting entities and relations from the scientific literature. It reduces error propagation between tasks and leverages cross-sentence relations through coreference links by introducing a multi-task setup and a coreference disambiguation task.

\noindent \textbf{DyGIE/DYGIE++} \cite{luan2019general, wadden2019entity}
dynamically build a span graph, and iteratively refine the span representations by propagating coreference and relation type confidences through the constructed span graph. Also, DyGIE++ takes event extraction into account.

\noindent \textbf{Multi-turn QA} \cite{li2019entity}
treats joint entity and relation extraction task as a multiple-round question-and-answer task. Each entity and each relation is depicted using a question-and-answer template, so that these entities and relations can be extracted by answering these templated questions.

\noindent \textbf{MRC4ERE++} \cite{zhaoasking} introduces a diversity question answering mechanism based on Multi-turn QA. Two answering selection strategies are designed to integrate different answers. Moreover, MRC4ERE++ proposes to predict a subset of potential relations to filter out irrelevant ones to generate questions effectively.

\subsection{Evaluation Metrics}
We evaluate these models on both entity recognition and relation extraction tasks. An entity is considered correct if its predicted span and entity label match the ground truth.
When evaluating relation extraction task, previous works have used different metrics. For the convenience of comparison, we report multiple evaluation metrics consistent with them. We define a \textbf{strict evaluation}, where a relation is considered correct if its relation type, as well as the two related entities, are both correct, and a \textbf{boundary evaluation}, where entity type correctness is not considered. We reported strict relation f1 on Conll04 and ADE, boundary relation f1 on SciERC, and both on ACE05. Our experiments on these datasets all report a micro-F1 score, except for the ADE dataset, where we report the macro-F1 score.

\subsection{Experiment Settings}

In most experiments, we use BERT  \cite{devlin2018bert}
as the encoder, pre-trained on an English corpus. On the SciERC dataset, we replace BERT with SciBERT  \cite{beltagy2019scibert}. We perform the four-level encoding with a subword encoding size $h=768$, a word encoding size $h_w=768$, a span encoding size $h_s=793$, and a span-pair encoding size $h_r=2354$. We set both entity memory slot size $h_{me}$ and relation memory slot size $h_{mr}$ to 768.  We just use a single graph neural layer in semantic and syntactic graphs. We initialize entity memory and relation memory using the normal distribution $\mathcal{N}(0.0,0.02)$.
We use the Adam Optimizer with a linear warmup-decay learning rate schedule (with a peak learning rate of 5e-5), a dropout before the entity and relation bilinear classifier with a rate of 0.5, a batch size of 8, span width embeddings of 25 dimensions and max span-size of 10. The training is divided into two stages with the first stage of 18 epochs, and the second stage of 12 epochs
\footnote{Our code will be available at \url{https://github.com/tricktreat/trimf}}.

\subsection{Results and Analysis}

\noindent\textbf{Main Results}
We report the average results over 5 runs on SciERC, ACE05 and CoNLL04 datasets. For ADE, we report metrics averaged across the 10 folds.
Table \ref{tab:results} illustrates the performance of the proposed method as well as baseline models on SciERC, ACE05, CoNLL04 and ADE datasets. Our model consistently outperforms the state-of-the-art models for both entity and relation extraction on all datasets. Specifically, the relation F1 scores of our model advance previous models by  +3.2\%, +4.9\%, +0.6\%, +2.3\% on SciERC, ACE05, CoNLL04 and ADE respectively. We attribute the improvement to three reasons. First, our model can share learned information between tasks through the Memory module, enhancing task interactions in both directions(from NER to RE, and from RE to NER). Second, the Trigger Sensor module can enhance the relation trigger information, which is essential for relation classification. Lastly, taking a step further from introducing structure information through syntactic graphs, we distinguish the semantic and syntactic importance of words to fuse two-way information through a dynamic Graph Weighted Fusion module. We conduct ablation studies to further investigate the effectiveness of these modules.

\subsection{Ablation Study}

\noindent\textbf{Effect of Different Modules}
To prove the effects of each proposed modules, we conduct the ablation study. As shown in Table \ref{tab:as_module}, all modules contribute to the final performance. Specifically, removing the Trigger Sensor module has the most significant effect, causing the relation F1 score to drop from 52.44\% to 51.23\% on SciERC, from 62.77\% to 61.60\% on ACE05. 
Comparing the effects of Memory-Flow Attention at subword-level and word-level on the two datasets, we find that the improvement of MFA at subword-level is more significant. We thus believe that fine-grained semantic information is more effective for relation extraction. The performance of the Syntactic-Semantic Graph Weighted Fusion module varies widely across datasets, achieving an improvement of 1.09\% on ACE05, but only 0.61\% on SciERC. This may be related to the different importance of syntactic information for relation extraction on different domains.
\begin{table}
  \small
  \begin{tabular}{@{}lccccc@{}}
  \toprule
 	& \multicolumn{2}{c}{Entity} & \multicolumn{2}{c}{Relation} \\ 
 	\cmidrule(lr){2-3} \cmidrule(lr){4-5}
 	Method & F1 & $\Delta$ & F1 & $\Delta$ \\ \midrule
 	\multicolumn{5}{c}{SciERC} \\ \midrule
 	TriMF & 70.17 & - & \textbf{52.44} & - \\
 	\quad w/o Graph Weighted Fusion & 70.12 & -0.05 & 51.83 & -0.61 \\ 
 	\quad w/o Trigger Sensor  & 70.19 & +0.02 & 51.23 & -1.21 \\ 
 	\quad w/o Subword-level MFA  & 70.11 & -0.06 & 51.27 & -1.17 \\
 	\quad w/o Token-level MFA  & \textbf{70.21} & +0.04 & 51.78 & -0.66 \\\midrule
  	\multicolumn{5}{c}{ACE05} \\ \midrule
 	TriMF & \textbf{87.61} & - & \textbf{62.77} & - \\
  	\quad w/o Graph Weighted Fusion & 87.55 & -0.06 & 61.68 & -1.09 \\
 	\quad w/o Trigger Sensor  & 87.45 & -0.16 & 61.60 & -1.17 \\
 	\quad w/o Subword-level MFA  & 87.09 & -0.52 & 61.68 & -1.09 \\
 	\quad w/o Token-level MFA  & 87.42 & -0.19 & 62.02 & -0.75 \\
   \bottomrule
 \end{tabular}
 \caption{Effect of Different Modules}
 \label{tab:as_module}
\end{table}

\noindent\textbf{Effect of Interaction Between Two Subtasks}
There is a mutual dependency between the entity recognition and relation extraction tasks. Our framework models this relationship through the Multi-level Memory Flow Attention module. Depending on the memory that the attention mechanism relies on, it can be divided into Relation-specific MFA and Entity-specific MFA. The Relation-specific MFA module enhances the relation-related information based on the relation memory, allowing the entity recognition task to utilize the information already captured in the relation extraction task, as does Entity-specific MFA. To verify that the Memory Flow Attention module can facilitate the interaction between entity recognition and relation extraction, we perform ablation studies, as shown in Table \ref{tab:as_interaction}. On ACE05 and SciERC, both Entity-specific MFA and Relation-specific MFA bring significant performance improvement. In addition, the Relation-specific MFA improves more compared with Entity-specific MFA. We think the reason may be that our model performs entity recognition first and then relation extraction. This order determines that information from entity recognition has been used by relation extraction, but the information from relation extraction is not fed back to entity recognition. When using Relation-specific MFA, a bridge for bi-directional information flow is built between the two tasks. Furthermore, when we use both Entity-specific MFA and Relation-specific MFA, the experiment achieves the best performance, indicating that MFA can enhance the bi-directional interaction between entity recognition and relation extraction.

\begin{table}
\small
  \begin{tabular}{@{}lccccc@{}}
  \toprule
 	& \multicolumn{2}{c}{Entity} & \multicolumn{2}{c}{Relation} \\ 
 	\cmidrule(lr){2-3} \cmidrule(lr){4-5}
 	Method & F1 & $\Delta$ & F1 & $\Delta$ \\ \midrule
 	\multicolumn{5}{c}{SciERC} \\ \midrule
  	TriMF  &  \textbf{70.17} & - &  \textbf{52.44} & -\\ 
 	\quad w/o MFA & 70.04 & -0.13 & 50.78 & -1.66\\
 	\quad w/o Relation MFA  & 70.07 & -0.10 & 51.28 & -1.16 \\
 	\quad w/o Entity MFA  & 70.17 & 0 & 51.84 & -0.60 \\ \midrule

  	\multicolumn{5}{c}{ACE05} \\ \midrule
   	TriMF  &  \textbf{87.61} & - &  \textbf{62.77} & -\\
  	\quad w/o MFA & 87.42 & -0.19 & 62.19 & -0.58\\
 	\quad w/o Relation MFA  & 87.37 & -0.24 & 62.06 & -0.71\\
 	\quad w/o Entity MFA  & 87.38  & -0.23 & 62.64 & -0.13 \\

   \bottomrule
 \end{tabular}
 \caption{Effect of Interaction between NER and RE}
  \label{tab:as_interaction}
\end{table}

\noindent\textbf{Effect of Different Graph Fusion Methods}
Our proposed graph weighted fusion module employs a node-wise weighted fusion approach based on attention, which enables a flexible fusion of node representations according to words' syntactic importance and semantic importance. To demonstrate the effectiveness of our approach, we compare other node-wise fusion methods, including no-fusion, max-fusion, mean-fusion and sum-fusion, as shown in Table \ref{tab:as_fusion}. Comparing the two experiments which only use the semantic graph or syntactic graph, we find that the syntactic graph provides a greater improvement in model performance, probably because the initial encodings of the nodes of the syntactic graph have already contained semantic information. Compared to max-fusion, mean-fusion, and sum-fusion, the node-wise weight-fusion method brings more improvement on relation F1 scores of both SciERC and ACE05, which proves the effectiveness of our method.

\begin{table}
\small
  \begin{tabular}{@{}lccccccc@{}}
  \toprule
 	& \multicolumn{3}{c}{Entity} & \multicolumn{3}{c}{Relation} \\ 
 	\cmidrule(lr){2-4} \cmidrule(lr){5-7}
 	Method & P & R & F1 & P & R & F1 \\ \midrule
 	\multicolumn{7}{c}{SciERC}\\ \midrule
 	No Graph & 69.87 & 70.33 & 70.10 & 52.56 & 49.59 & 51.03 \\
 	Semantic Graph & 68.47 & 69.61 & 49.04 & 52.00 & 50.62 & 51.30 \\
 	Syntactic Graph & 72.18 & 70.68 & \textbf{71.42} & 54.02 & 48.97 & 51.37  \\
 	Mean-fusion & 69.77 & 69.02 & 69.39 & 53.56 & 49.39 & 51.38 \\
 	Sum-fusion & 69.45 & 69.57 & 69.51 & 52.94 & 49.65 & 51.24 \\
 	Max-fusion & 69.12 & 69.64 & 69.38 & 53.01 & 49.45 & 51.17 \\
 	Weighted fusion & 70.18 & 70.17 & 70.17 & 52.63 & 52.32 & \textbf{52.44} \\ \midrule
 	\multicolumn{7}{c}{ACE05}\\ \midrule
 	No Graph & 87.24 & 87.18 & 87.21 & 60.11 & 61.83 & 60.96 \\
 	Semantic Graph & 87.57 & 87.69 & 87.63 & 59.45 & 62.47 & 60.92 \\
 	Syntactic Graph & 87.47 & 87.36 & 87.41 & 59.29 & 62.96 & 61.07  \\
    Mean-fusion & 87.32 & 87.78 & 87.55 & 59.74 & 62.90 & 61.28 \\
 	Sum-fusion & 87.85 & 87.47 & \textbf{87.66} & 60.12 & 62.26 & 61.17 \\
 	Max-fusion & 87.51 & 87.62 & 87.56 & 60.22 & 62.25 & 61.22 \\
 	Weighted fusion & 87.67 & 87.54 & 87.61 & 62.19 & 63.37 & \textbf{62.77} \\
   \bottomrule
 \end{tabular}
 \caption{Effect of Different Graph Fusion Methods}
 \label{tab:as_fusion}
\end{table}

\begin{table*}
\small
  \begin{tabular}{@{}m{7.8cm}m{4cm}c@{}}
  \toprule
 	Original Text & Relation & Top-5 Relation Triggers \\ \midrule
 	Urutigoechea and the others were arrested Wednesday in the cities of Bayonee and \textbf{Bonloc} in southwestern \textbf{France} in Poitiers in west-central France.  & (Bonloc, Located in, France) & southwestern, west-central, cities, of, in  \\
\textbf{Kleber Elias Gia Bustamante}, accused by the police of being a member of the "Red Sun" central committee, has been living clandestinely since his escape from the Garcia Moreno Prison, where he was held accused of assassinating the industrialist, \textbf{Jose Antonio Briz Lopez}.  & (Kleber Elias Gia Bustamante, Kill, Jose Antonio Briz Lopez) & Prison, assassinating, held, of, accused\\
   \bottomrule
 \end{tabular}
 \caption{Results of Trigger Words Extraction}
 \label{tab:trigger_extraction}
\end{table*}

\begin{figure}[h]
  \centering
  \includegraphics[width=\linewidth]{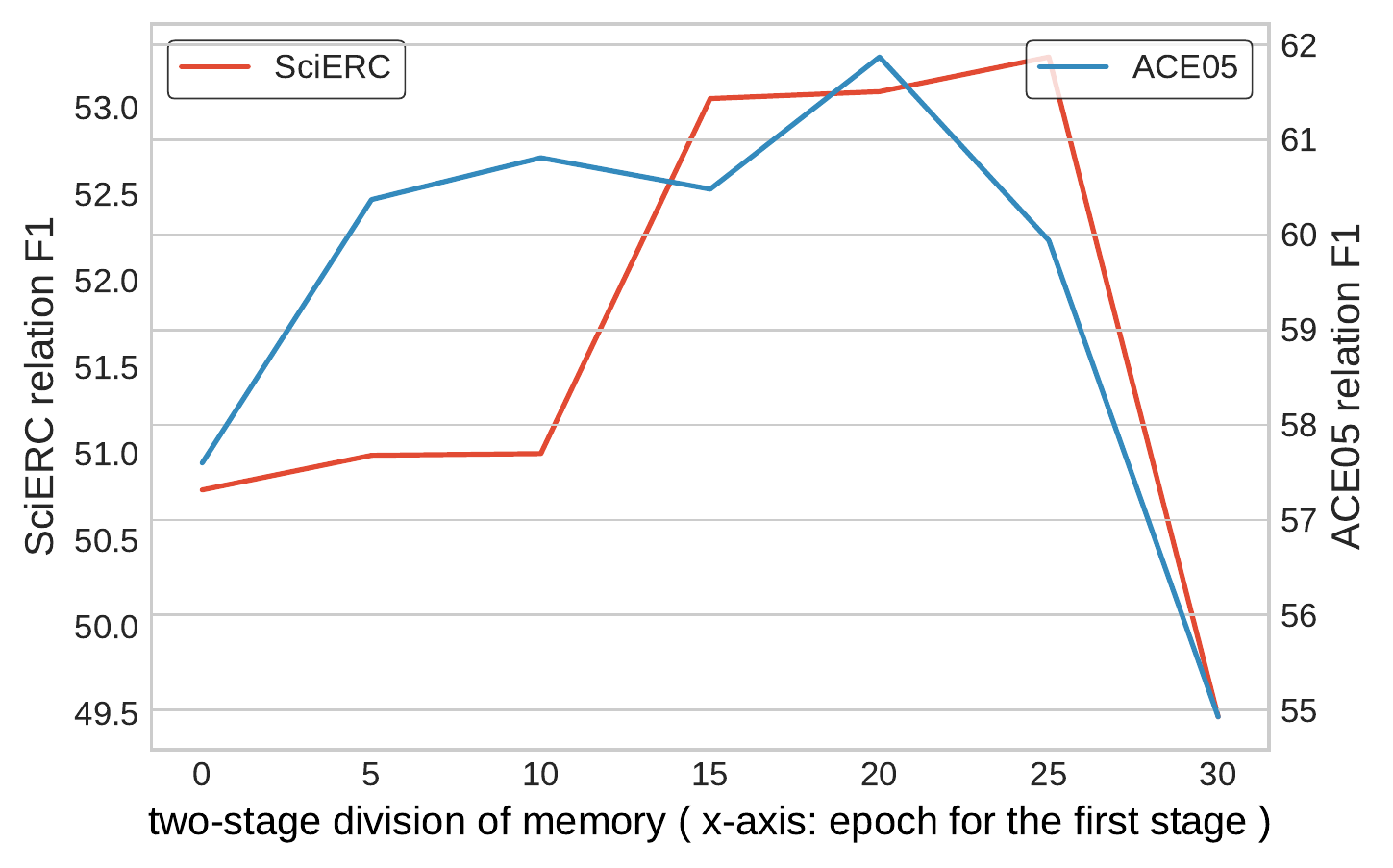}
  \caption{Effect of Train Stage Division}
   \label{fig:stage_division}
\end{figure}

\noindent\textbf{Effect of Different Stage Divisions for Memory}
We explored the effect of different two-stage divisions on the relation classification, as shown in Figure \ref{fig:stage_division} (x-axis is the number of epochs for the first stage and the total number of epochs is 30). We can note that if our model skips the first stage (x=0) or ignores the second stage (x=30), the performance of the model degrades significantly. Specifically, as the proportion of first stage epochs to total epochs increases, our model performs better. But at a certain point, the performance degrades significantly. We believe this is due to a decrease in epochs of the second stage and the memory already written in the first stage is not utilized effectively. Therefore the two-stage training strategy is effective, and a good balance of the two stages can bring out a better model performance. 

\noindent\textbf{Effect of Different Gradients Flow to Memory}
Our model primarily writes the memory in Memory-Aware Classifier. Furthermore, we can also tune the memory in MFA and Trigger Sensor modules through the backpropagation of gradients. The gradient flows are divided into three types: Trigger Sensor gradients, Subword-level MFA gradients and  Word-level MFA gradients, and we investigated the effects of different gradients, as shown in Table \ref{tab:as_grad}. We see that on the ACE05 dataset, when we block any of the gradients flows, the model performance decreases significantly, by 1.35\%, 1.54\%, and 0.92\% on relation F1 score, which indicates that tuning the memory during the second stage is effective. However, On the SciERC dataset, there is no significant drop, and we believe that the model has learned accurate representations of the categories in the first training stage.

\begin{table}
\small

  \begin{tabular}{@{}lcccccc@{}}
  \toprule
 	& \multicolumn{2}{c}{Entity} & \multicolumn{2}{c}{Relation} \\ 
 	\cmidrule(lr){2-3} \cmidrule(lr){4-5}
 	Method & F1 & $\Delta$ & F1 & $\Delta$ \\ \midrule
 	\multicolumn{5}{c}{SciERC} \\ \midrule
 	TriMF  & 70.17 & - & \textbf{52.44} & - \\
 	 \quad w/o Trigger Sensor Grad.  & 70.14 & -0.03 & 52.28 & -0.16 \\
 	 \quad w/o Subword-level MFA Grad.  & \textbf{70.23} & +0.08 & 52.03 & -0.41 \\
 	 \quad w/o Word-level MFA Grad.  & 70.12 & -0.05 & 52.14 & -0.30\\ \midrule
  	\multicolumn{5}{c}{ACE05} \\ \midrule
 	TriMF  & \textbf{87.61} & - & \textbf{62.77} & - \\
 	 \quad w/o Trigger Sensor Grad.  & 87.55 & -0.06 & 61.42 & -1.35 \\
 	 \quad w/o Subword-level MFA Grad.  & 87.43 & -0.18 & 61.23 & -1.54 \\
 	\quad w/o Word-level MFA Grad.  & 87.34 & -0.27 & 61.85 & -0.92\\
   \bottomrule
 \end{tabular}
 \caption{Effect of Gradient Flow to Memory}
   \label{tab:as_grad}
\end{table}

\begin{figure}[h]
  \centering
  \includegraphics[width=\linewidth]{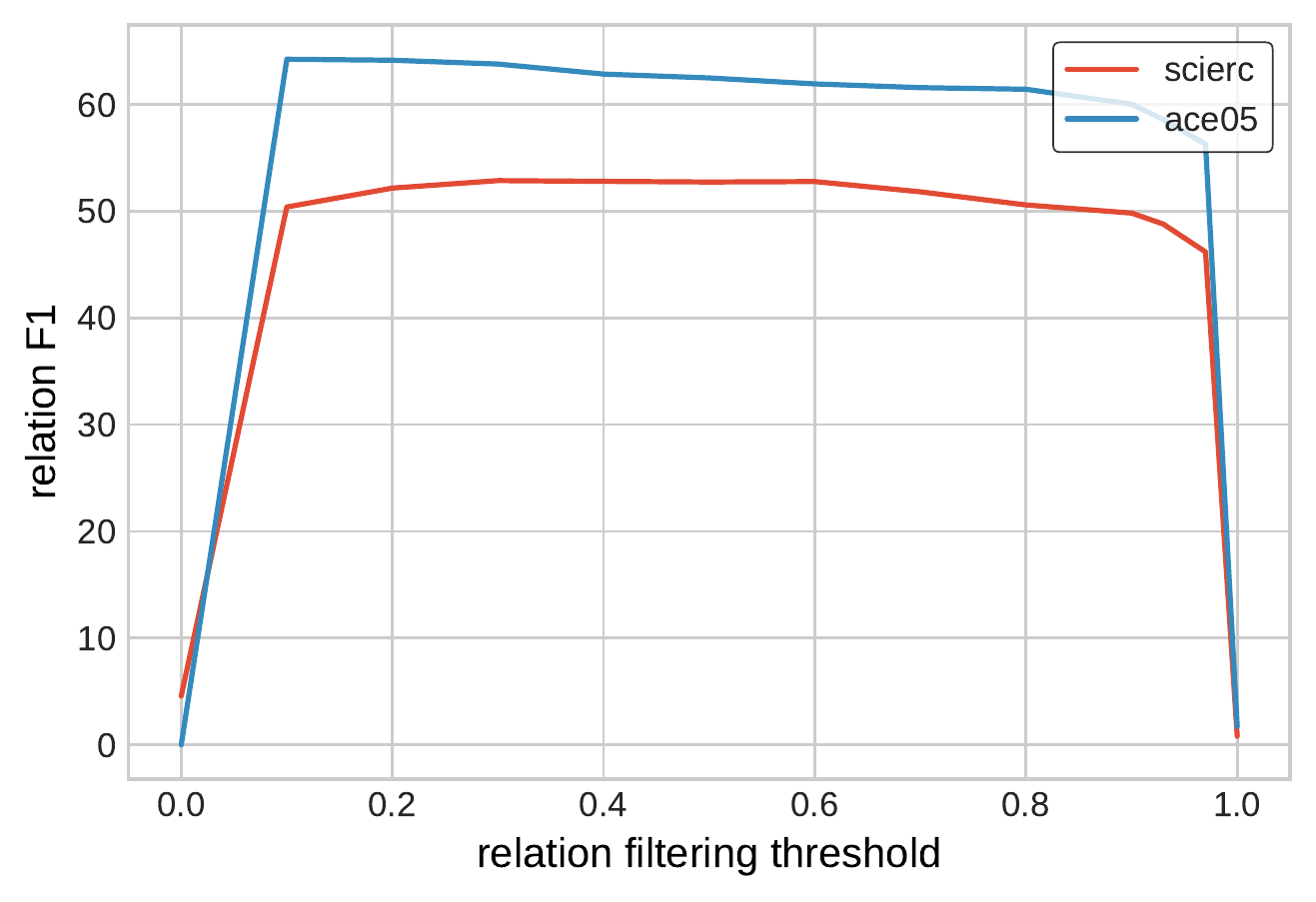}
  \caption{Effect of Relation Filtering Threshold}
  \label{fig:figure_threshold}
\end{figure}

\noindent\textbf{Effect of Relation Filtering Threshold}
The precision and recall of relation classification are correlated with predefined thresholds. We investigate the impact of the relation filtering threshold on relation F1. Figure  \ref{fig:figure_threshold} shows the relation F1 score on the SciERC and ACE05 test sets, plotted against the relation filtering threshold. We see that the performance of our model is stable for the choice of relation filtering thresholds. Our model is able to achieve good results on relation classification except for extreme thresholds of 0.0 or 1.0. Therefore, within a reasonable range, our model is not sensitive to choose a threshold.

\subsection{Case Study}

\begin{figure*}[h]
  \centering
  \includegraphics[width=0.9\linewidth]{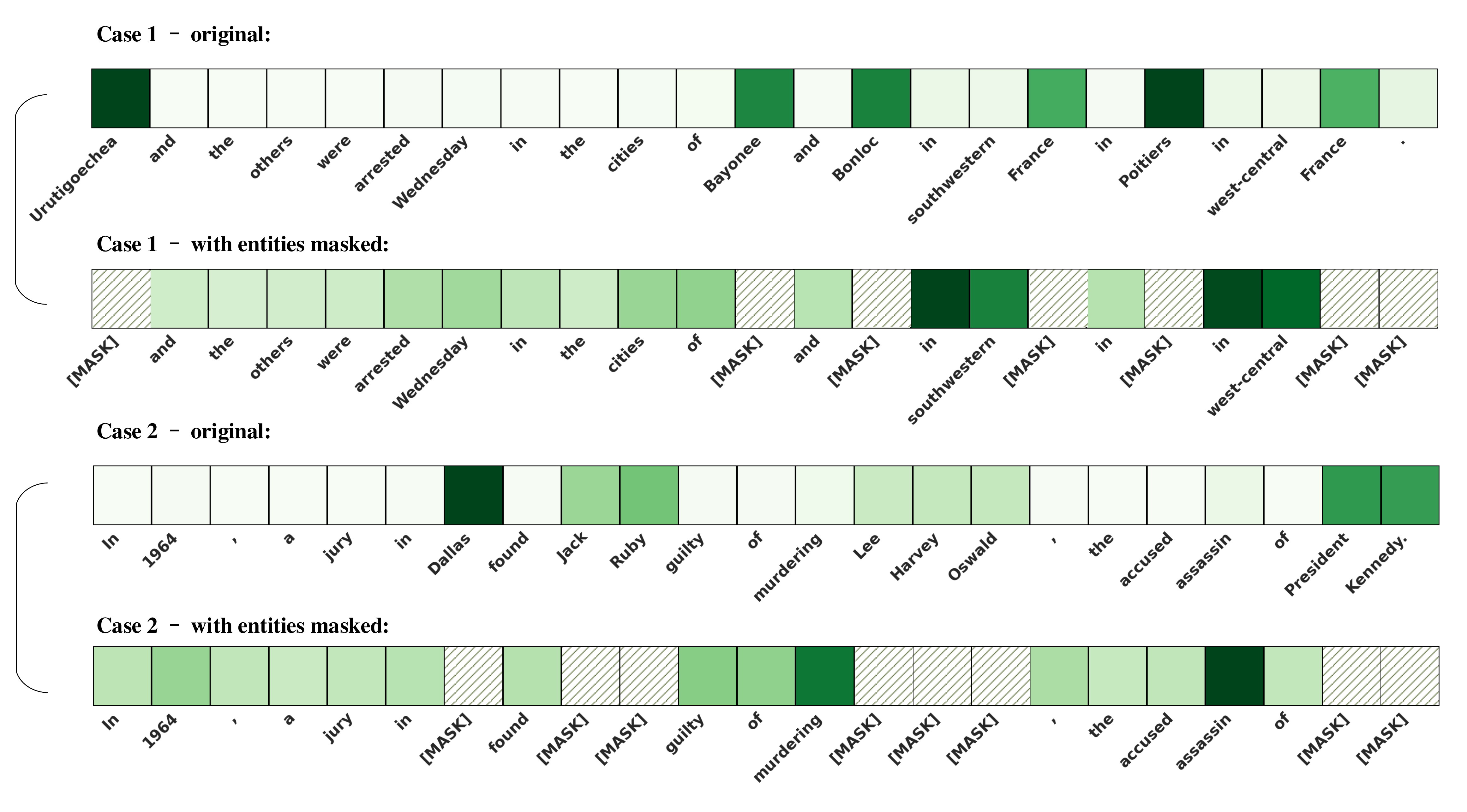}
  \caption{Two case studies of relation memory flow attention during inference. The darker cells have higher attention weights.}
  \label{fig:vis_attn}
\end{figure*}

\begin{table*}
\small
  \begin{tabular}{@{}m{17.7cm}@{}}
  \toprule
  
    The problem is not unusual in {\color{red}[}{\color{blue}[} Guernsey {\color{blue}]$_{\text{H\_Located-in}}$}{\color{red}]$_{\text{H\_Located-in}}$}, one of {\color{blue}[} Britain {\color{blue}]$_{\text{T\_Located-in}}$} 's {\color{blue}[} Channel Islands {\color{blue}]$_{\text{T\_Located-in}}$} off the coast of {\color{red}[}{\color{blue}[} France {\color{blue}]}{\color{red}]$_{\text{T\_Located-in}}$} \\
\midrule

... and former  {\color{red}[}{\color{blue}[} CBS {\color{red}]$_{\text{T\_Work-for}}$} News {\color{blue}]} commentator  {\color{red}[}{\color{blue}[} Eric Sevareid {\color{blue}]$_{\text{H\_Live-in}}$} {\color{red}]$_{\text{H\_Work-for}}$}  , who was born in {\color{blue}[} Velva {\color{blue}]$_{\text{T\_Live-in}}$}  , several miles southeast of {\color{blue}[} Minot {\color{blue}]}. \\\midrule

  Text of the statement issued by the {\color{blue}[} Organization of the Oppressed on Earth {\color{blue}]} claiming {\color{red}[}{\color{blue}[} U. S.  {\color{blue}]$_{\text{T\_Live-in}}$}{\color{red}]$_{\text{T\_Live-in}}$} Marine Lt.{\color{red}[}{\color{blue}[} William R. Higgins {\color{blue}]$_{\text{H\_Live-in}}$}{\color{red}]$_{\text{H\_Live-in}}$} was hanged.\\ 

   \bottomrule
 \end{tabular}
 \caption{Typical error examples. Red brackets indicate entities predicted by the model, blue brackets indicate true entities, and the labels in the lower right corner indicate the type of relation  between the corresponding entities and the head or tail type (T for the tail entity; H for the head entity)}
   \label{tab:error}
\end{table*}

\noindent\textbf{Trigger Words Extraction}
With the Trigger Sensor module, our model has the ability to extract the relation triggers. We rank the similarities of each word representation with the span-pair specific relation representation, which have been calculated in the Trigger Sensor. Filtering out the entity surface words and stopwords, the top k words are picked as relation triggers and used to interpret the results of the relation extraction. We show two cases in Table \ref{tab:trigger_extraction}.

\noindent \textbf{Memory Flow Attention Visualization}
We visualize the weights of attention to provide a straightforward picture of how the entity and relation memory flow attention we designed can both enhance the interaction between entity recognition and relation extraction. Also, it can enhance the information about relation triggers in context, to some extent explaining the model's predictions.
Figure \ref{fig:vis_attn} shows two cases of how attention weights on context from a relation memory flow can help the model recognize entities and highlight relation triggers. Each example is split into two visualizations, with the top showing the original attention weights and the bottom showing the attention weights after masking the entities. In the top figure, we can see that the darker words belong to an entity, for example, \textit{"Urutigoechea"}, \textit{"Bayonee"}, \textit{"Bonloc"} in case 1, \textit{"Dallas"}, \textit{"Jack Ruby"} in case 2,  illustrating that the attention of our relation memory flow attention can highlight relevant entity information. Consistent with \cite{han2020more}, our attention distribution also illustrates that entity names provide more valid information for relation classification compared to context. To more clearly visualize the attention weights of different contextual words, we mask all entities, formalize the weights of the remaining words, and then visualize them. As shown in the bottom figure, the relation memory flow pays more attention on the words that indicate the type of relation, i.e., relation triggers, such as \textit{"in"}, \textit{"southwestern"}, \textit{"west-central"} in case 1 can indicate \textit{"Located in"} relation, and \textit{"assassin"}, \textit{"murdering"} in case 2 can indicate \textit{"Kill"} relation. This shows that our relation memory flow is able to highlight relation triggers, helping the model with better performance on relation extraction.

\noindent\textbf{Error Cases} 
In addition to visualizing Memory Flow Attention weights on true positives, we also analyze a number of \textbf{false positives} and \textbf{false negatives}. These error cases include relation requiring inference, ambiguous entity recognition and long entity recognition, as shown in Table \ref{tab:error}. In the first case, although our model is able to recognize the four entities about \textit{Location}, it incorrectly extracts the relation \textit{"(Guernsey, Located in, France)"} and does not extract the correct one \textit{"(Guernsey, Located in, Channel Islands)"}. This is because the model does not infer the complex location relation between the four entities. Our model is prone to make mistakes when classifying ambiguous entities, and False Positive and False Negative often occur together. For example, in the second row of the Table \ref{tab:error}, the model does not recognize \textit{"CBS News"} as a Location entity, but recognizes \textit{"CBS"} which is not labeled in the test set.
Furthermore, recognition of long entities is a challenge for our model due to the fact that long entities are sparse in the dataset. For example, in the third row of the Table \ref{tab:error}, the model fails to recognize the long entity \textit{"Organization of the Oppressed on Earth"}.

\section{Conclusion and Future Work}

In this paper, we propose a Trigger-Sense Memory Flow Framework (TriMF) for joint entity and relation extraction. We use the memory to boost the task-related information in a sentence through the Multi-level Memory Flow Attention module. This module can effectively exploit the mutual dependency and enhance the bi-directional interaction between entity recognition and relation extraction tasks. Also, focusing on the relation triggers, we design a Trigger Sensor to sense and enhance triggers based on memory. Our model can extract the relation triggers without any trigger annotations, which can better assist the relation extraction and provide an explanation. Furthermore, we distinguish the semantic and syntactic importance of a word in a sentence and fuse semantic and syntactic graphs dynamically based on the attention mechanism. Experiments on SciERC, ACE05, CoNLL04 and ADE datasets show that our proposed model TriMF achieves state-of-the-art performance.

In the future, we will improve our work along with two directions. First, we plan to impose constraints on the representations of entity categories and relation categories written in the memory, due to the fact that relations and entities substantively satisfy specific constraints at the ontology level. Second, for improving the model's ability on sensing the trigger, we plan to add weak supervision (e.g. word frequency, entity boundary) to the Trigger Sensor module.

\begin{acks}
This work is supported by the National Key Research and Development Project of China (No. 2018AAA0101900), the Fundamental Research
Funds for the Central Universities, the Chinese Knowledge Center of Engineering Science and Technology (CKCEST) and MOE Engineering Research Center of Digital Library.

\end{acks}

\bibliographystyle{ACM-Reference-Format}
\bibliography{main}

\end{document}